\documentclass[sigconf]{acmart}

\usepackage{booktabs} 
\usepackage{todonotes}
\usepackage{url}
\usepackage{algorithm,algorithmic}
\usepackage{subfigure}
\usepackage{balance}

\setcopyright{acmcopyright}

\renewcommand\footnotetextcopyrightpermission[1]{} 
\title{EdgeNet - Balancing Accuracy and Performance for Edge-based $\newline$ Convolutional Neural Network Object Detectors}

\begin{document}
\copyrightyear{2019} 
\acmYear{2019 \newline} 
\acmConference[ICDSC 2019]{13th International Conference on Distributed Smart Cameras}{September 9--11, 2019}{Trento, Italy}
\acmPrice{15.00}
\acmDOI{10.1145/3349801.3349809}
\acmISBN{978-1-4503-7189-6/19/09}

\author{George Plastiras}
\affiliation{%
  \institution{Department of Electrical and Computer Engineering,\\KIOS Research and Innovation Center of Excellence,\\University of Cyprus}
  \streetaddress{P.O. Box 1212}
  \city{Nicosia, Cyprus}
}
\email{gplast01@ucy.ac.cy}

\author{Christos Kyrkou}
\orcid{0000-0002-7926-7642}
\affiliation{%
  \institution{KIOS Research and Innovation Center of Excellence,\\University of Cyprus}
  \streetaddress{P.O. Box 1212}
  \city{Nicosia, Cyprus}
}
\email{kyrkou.christos@ucy.ac.cy}

\author{Theocharis Theocharides}
\affiliation{%
  \institution{Department of Electrical and Computer Engineering,\\KIOS Research and Innovation Center of Excellence,\\University of Cyprus}
  \streetaddress{1 Th{\o}rv{\"a}ld Circle}
  \city{Nicosia, Cyprus}
}
\email{ttheocharides@ucy.ac.cy}


\begin{abstract}

Visual intelligence at the edge is becoming a growing necessity for low latency applications and situations where real-time decision is vital. Object detection, the first step in visual data analytics, has enjoyed significant improvements in terms of state-of-the-art accuracy due to the emergence of Convolutional Neural Networks (CNNs) and Deep Learning. However, such complex paradigms intrude increasing computational demands and hence prevent their deployment on resource-constrained devices. In this work, we propose a hierarchical framework that enables to detect objects in high-resolution video frames, and maintain the accuracy of state-of-the-art CNN-based object detectors while outperforming existing works in terms of processing speed when targeting a low-power embedded processor using an intelligent data reduction mechanism. Moreover, a use-case for pedestrian detection from Unmanned-Areal-Vehicle (UAV) is presented showing the impact that the proposed approach has on sensitivity, average processing time and power consumption when is implemented on different platforms. Using the proposed selection process our framework manages to reduce the processed data by $~100\times$ leading to under $4W$ power consumption on different edge devices.

\end{abstract}



\maketitle

\colorlet{hl}{green!10!orange!90!}

\section{Introduction}
Visual intelligence is a rapidly growing field that can provide improved high-level understanding of the environment. Computer vision algorithms, in particular, are increasingly employed on mobile/edge devices that support high-resolution cameras. Applications such as emergency response, disaster management, and recovery, and monitoring of critical infrastructures, can all benefit from real-time video analytics. In many cases, for such applications the connectivity to a cloud service may not be available or not existent at all. Furthermore, processing information on-board can eliminate security issues when transmitting sensitive information for such applications. Hence, on-board processing is highly desirable at the edge. 

In particular, object detection, the first step in visual data analytics, has recently enjoyed significant accuracy and performance improvements due to the emergence of deep learning and the technology advances in Graphical Processing Units(GPU), respectively. However, such complex paradigms intrude increasing computational demands and are not traditionally implemented in resource-constrained devices. 

Convolutional Neural Networks (CNNs) build hierarchical representations that can efficiently perform a variety of vision tasks such as detection, recognition and segmentation \cite{6296526}, \cite{DBLP:journals/corr/HeGDG17}. To facilitate the mapping of CNNs on resource constrained devices, recent works have focused on co-designing for high task-level accuracy and low computational complexity. This has been addressed from different aspects, with emphasis on precision reduction, network pruning, and compression as well as compact network design. Furthermore, such optimizations works mostly on small and fixed image size and do not consider applications, such as Unmanned Aerial Vehicles (UAV) that need to work on higher resolution images. As such there is still a need to accommodate improvements in CNN architectures and design techniques with intelligent data reduction to maximize the efficiency of CNNs for such applications. 

\begin{figure}[t]
\centering
\includegraphics[width=1\linewidth]{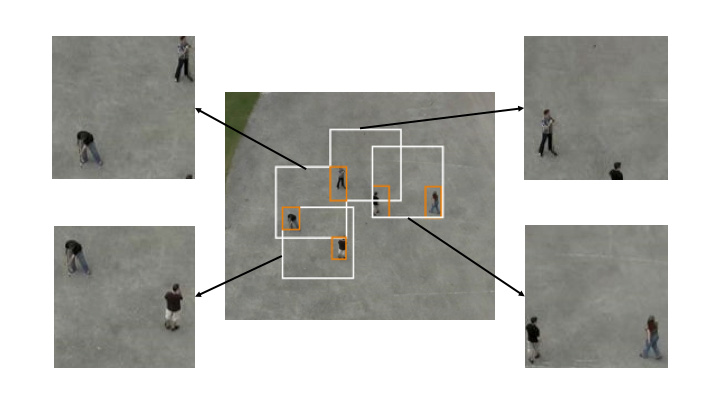}
\vspace{-10mm}
\caption{Proposed tiles for processing base on the selection process.}
\label{tile}
\vspace{-6mm}
\end{figure}

Thus, our contribution focuses on an intelligent way to reduce the processed data by using the proposed $EdgeNet$ framework, that can work with any predefined architecture, on larger scale images leading to an increase of both accuracy and overall performance of the system. We propose a way of focusing only on promising regions and examine the impact of building resolution-optimized networks to further improve the computation and accuracy trade-offs, as shown in Fig. \ref{tile}. 


$EdgeNet$ framework consist of three main stages: 
\begin{itemize}
    \item An optimized CNN, called $DroNet\_V3$ that is lightweight and operates on lower resolution input to provide initial estimates for object positions
    \item A pool of per-scale-  and region-size- optimized CNNs, called $DroNet\_Tile$ out of which the most suitable for processing are selected at each time instance based on statistical metrics
    \item An optical-flow tracker to compensate for the increasing demands of the previous stages and speed-up of the whole process
\end{itemize}
The proposed framework was evaluated and compared with state-of-the-art object detectors using a pedestrian dataset from Unmanned Aerial Vehicle(UAV)-captured images, on an i5 CPU and two ARM-based CPUs on different platforms. Throughout the analysis on our test dataset, we demonstrate that the detection accuracy can considerably improve $~6-20\%$, along with a $~1.5\times$ reduction on the energy consumption of the system while increasing the performance $~60-100 \times$ compared to state-of-the-art CNNs. EdgeNet, is able to maintain the accuracy of a high-end implementation, while outperforming existing works in terms of processing speed and energy consumption when targeting a low-power embedded processor implementation, without changing the structure of an existing network just by intelligently selecting regions of the image.

\begin{figure*}[h]
\centering
\includegraphics[width=1\linewidth]{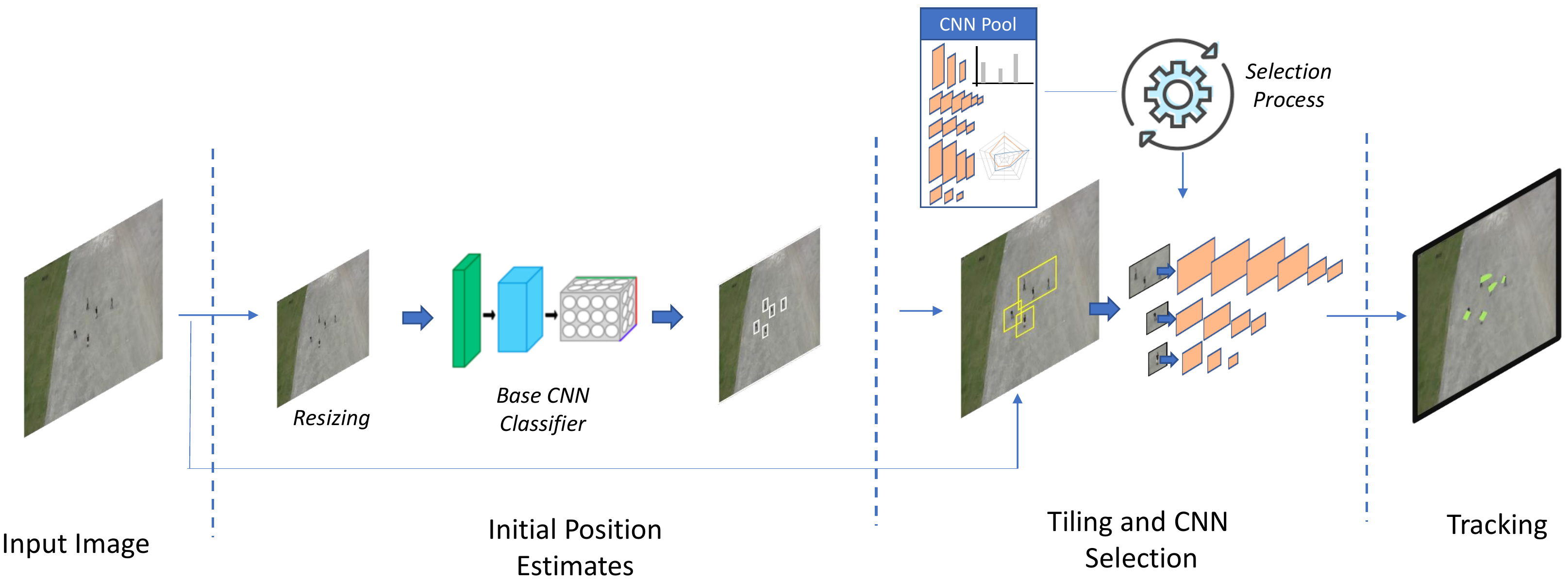}
\caption{Overview of $EdgeNet$}
\label{framework_overview}
\end{figure*}


\section{CNN inference at the Edge}

Convolutional Neural Networks have shown remarkable promise in a variety of scenarios with impressive accuracy and performance. In most cases, this comes at the cost of high computational, power and memory requirements. In typical application scenarios, these CNNs run on powerful GPUs that consume a lot of power. In response to the excessive resource demands of CNNs, the traditional way is to use powerful cloud datacenter for training and evaluating CNNs \cite{8117585}. Input data generated from mobile devices are sent to the cloud for processing, and the results are sent back to the mobile devices after the inference. This cannot be applied in some cases such as search and rescue missions in remote areas or in cases of natural disaster where the network grid might not be available.
With the advancement of the technology and the powerful devices such as Jetson by Nvidia\cite{Jetson} and Edge TPU by Google\cite{TPU}, that can analyze real-time data at the edge, a new wealth of possibilities opens up for potential applications, including sensing the user's immediate environment, navigating, assisting medical professionals, and home automation \cite{Chinta2014ANA}, \cite{738305}. 

However, in some cases particularly for high resolution image processing, the use of deep neural networks on devices like mobile phones or smart watches is challenging, since model sizes are large and do not fit in the limited memory available on such devices. Recent works are looking to minimize the size of the neural networks, while maintaining accuracy, using different strategies such as down-sampling and filter count reduction \cite{2017arXiv170404861H},\cite{2016arXiv160207360I}. Other works \cite{ML},\cite{2017arXiv170800630R}, focus on creating specialized frameworks to compress the neural network models, using state-of-the-art techniques such as pruning on weights and operations that are least useful for prediction, quantization by reducing the number of bits for model weights and activations. 


Other approaches look at the optimization beyond the CNN optimization \cite{FRCNN2017}, \cite{DBLP:journals/corr/ErhanSTA13}. These CNNs are working on a region proposal base, where they use a small network to slide over a convolutional feature map in order to generate proposal for the region where the object lies. Different anchor boxes are proposed for each position of an image in order to be examined by a classifier and regressor to check the occurrence of objects. On the other hand, some approach are trying to look at the image only once \cite{YOLOv2},\cite{DBLP:journals/corr/LiuAESR15}, and predict this boxes without the two stage approach of the region proposal and the large amount of proposals that need to process. Moreover, in \cite{Plastiras:2018:ECO:3243394.3243692}, we presented a Selective Tile Processing approach where instead of resizing the input image and process it with a CNN, we selected only regions of the image for processing in a static separation of the input image on same sized tiles, based on the image and CNN input.

To this end, in this work we focus on techniques beyond the CNN optimization, in order to intelligently reduce the data that need to be processed by a CNN and enable real-time processing on mobile/edge devices on high resolution images. In particular, we focus on techniques that reduce both the large amount of proposals of Region Proposal networks, and the times an image is resized on Single-Shot networks. Based on the Selective Tile Processing approach \cite{Plastiras:2018:ECO:3243394.3243692}, we proposed a framework that evaluates and dynamically select regions of the image based on statistical metrics gathered from previous frames. In particular, we are able to use smaller structures of CNN that can utilize efficiently the tiling approach and avoid the static separation of the input image.

\section{Proposed Approach}\label{approach}
We propose $EdgeNet$, a framework based on multiple CNN detectors aiming to improve the overall performance of both accuracy and processing time along with a reduction of power consumption, of an edge-based detector on high-resolution images. Moreover, we present an evaluation of different algorithmic parameters and configurations, an indication of the number of frames the framework must spend at each stage before moving to the next stage, in order to analyze the impact on both performance and accuracy of the detections. To this end, Fig. \ref{framework_overview} shows the pipeline of $EdgeNet$ framework, which consist of three-stages. A detailed description of each stage is given below:

\begin{table}[]
\begin{tabular}{|l|c|c|}
\hline
\textbf{CNN} & \multicolumn{1}{l|}{\textbf{Input Size (\textit{pixels})}} & \multicolumn{1}{l|}{\textbf{Processing Time (\textit{sec})}} \\ \hline
\textit{DroNet\_V3} & 512 & 0.08 \\
\textit{DroNet\_Tile} & 512 & 0.03 \\
\textit{DroNet\_Tile} & 416 & 0.02 \\
\textit{DroNet\_Tile} & 352 & 0.014 \\
\textit{DroNet\_Tile} & 256 & 0.008 \\
\textit{DroNet\_Tile} & 128 & 0.002 \\ \hline
\end{tabular}
\caption{Procesing time of $DroNet\_V3$ and $DroNet\_Tile$ for different input sizes indicating the Pool of CNNs}
\label{cnn_sizes}
\end{table}


\textbf{$\textbf{Initial Position Estimation}:$} The first stage of our framework is responsible for producing the initial positions of objects in a frame, thus an appropriate method must be selected that is accurate enough to steer the framework in the right direction. For this task, we used an efficient Convolutional Neural Network designed for edge applications\cite{DroNet2018}. We extend the structure of this network by up-sampling feature maps from earlier layers to detect object at multiple scales leading to a sufficient improvement on the accuracy of the detector \cite{DBLP:journals/corr/abs-1804-02767} for smaller objects, such as pedestrians, that we are going to refer to as $DroNet\_V3$. This stage works with the traditional way of resizing the input image, passed it through the $DroNet\_V3$ and then the produced detections are saved as a set of bounding boxes where each box correspond to an object in the image.

\textbf{$\textbf{Tiling and CNN Selection}:$} The second stage of the proposed framework is responsible to reduce the data that need to be processed by a CNN detector. The idea is to select different regions of the image, referred to as \textit{tiles}, to find the minimum image region that needs to be processed, based on the detected positions of the objects in prior time instances. To be able to illustrate the whole process, we are going to use the proposed CNN \cite{DroNet2018} that operates on different input sizes depending on the tile size ($128 - 512$) and refer to it as $DroNet\_Tile$.

Prior to the selection process it is necessary to perform a profiling and benchmarking of CNNs with different input sizes, between $128 - 512$ in our case as shown in Table \ref{cnn_sizes}. These CNNs make up a pool out of which the best ones will be chosen at every time instance to guarantee the minimum processing time. 

\begin{figure}[h]
\includegraphics[width=1\linewidth]{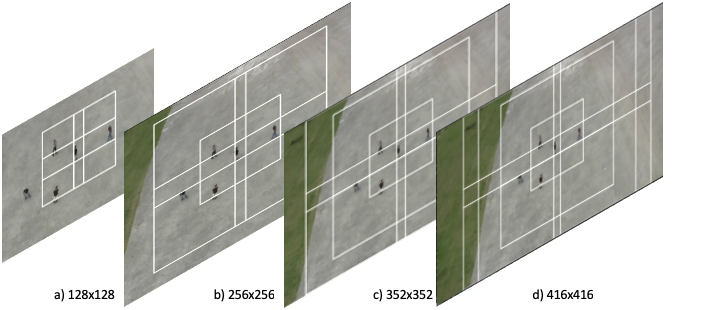}
\caption{ Different tile proposals, with respect to the position and size of the tiles, for an object in the image. a) $128\times128$, b) $256\times256$, c) $352\times352$, d) $416\times416$}
\label{pro}
\end{figure}

In addition, we also utilize the number of objects in that tile as a factor to guide the selection. This procedure requires to identify candidate tiles that cover each object. Hence, for each detected box proposed by the first stage (Fig. \ref{framework_overview}) a number of tiles are generated by positioning the object at each of the four tile corners, as shown in Fig.\ref{pro}. In addition, tiles with different sizes are also generated, in our case we used a total of $5$ sizes: $512, 416, 352, 256, 128$ matching the different sizes in the CNN pool, as shown in Table \ref{cnn_sizes}. A total of $20$ tiles for each object are proposed, where each tile is evaluated by the selection process based on the objects that it covers and its associated processing time. Thus, for each of the $20$ tiles per object proposed we calculate an Effective Processing Time (EPT), which is the number of objects that are covered divided by its corresponding processing time (Table \ref{cnn_sizes}). From the proposed tiles per object we select the one with the minimum EPT. Finally, we combine all the extracted tiles for all objects, and discards the redundant ones (i.e., those that cover the same or fewer objects) and retain only the one with the minimum $EPT$.

For the example in Fig. \ref{tile} four $128\times128$ tiles are selected by the selection process. Each tile that is selected is processed by the appropriate CNN from the the pool, based on its size. To this end, the processing time will be $4\times0.002 = 0.008s$ using the selected tiles compared to $0.05s$ using the $DroNet_V3$, which shows a significant impact on the performance even on this simple example. 


\textbf{$\textbf{Optical flow based tracker}:$} The third and final stage of the proposed framework is an optical flow tracker, named Lucas-Kanade \cite{Lucas:1981:IIR:1623264.1623280}. Lucas-Kanade tracker works on the principle that the motion of objects in two consecutive images is approximately constant relative to the given object. The selection of this tracker was based on its fast execution time, even with a large number of tracked points in the image. It is worth mentioning that any other tracker will also work on this stage with regards to application requirements such as accuracy and speed trade-off.
This stage is used for two main reasons. 1) To track the objects of the framework along with stage $1$ and $2$ and compare and verify the position of the detected object using both tracking and detection algorithms and 2) to reduce the processing time of the framework using only the tracker, before detecting the whole image again. To be able to use this tracker, a centered point must be calculated for each detected box in the frame, based on stage $1$. These points are used along with the corresponding frame as the initialized points of the tracker. Each time the tracker is called, it uses the previous and current frame in order to calculate the optical flow of the points that correspond to the objects and returns the estimated new position of each object. Based on the application requirements and the processing platform, for having a good trade-off between accuracy and performance, a specific time-slot combination is selected, which determines how many times each stage will be executed in the process loop as described in Section \ref{analysis}.


\section{Training Dataset For UAV Case Study}
Images were collected using manually annotated video footage from a UAV and the UCF Aerial Action Data Set \cite{UCFAerialActionDataSet} in order to train $DroNet\_Tile$, $DroNet\_V3$ and $Tiny-YoloV3$ \cite{DBLP:journals/corr/abs-1804-02767} to detect pedestrians in a variety of scenarios, and different conditions with regards to illumination, viewpoint, occlusion, and backgrounds. Overall, for the training set a total of $1500$ images were collected with a total of $60000$ pedestrians captured. We used Darknet \cite{darknet13}, a C- and CUDA-based Neural Network framework, to train, test and evaluate each CNN on different platforms. Each CNN that we tested was trained on the Titan Xp GPU for $200000$ iterations on the same dataset.

\section{Evaluation and Experimental results}
In this section, we present an extensive evaluation of the proposed $EdgeNet$ framework for different configurations. The configurations differ in the amount of time (number of frames) that is allocated to each stage. Specifically we use the notation $N_{S1}-N_{S2}-N_{S3}$, to indicate the number of frames that $EdgeNet$ affords to each stage. Moreover, we present an extensive evaluation and comparison with three different single-shot models $DroNet\_Tile$ , $DroNet\_V3$, and $Tiny-YoloV3$ that vary in terms of computational complexity. In this way we demonstrate that any approach not utilizing some form of tiling and dynamic selection exhibits accuracy drop because of the reduced image resolution. We also compare each of them for different three different computational platforms that facilitate different use-cases. The CNNs were trained and tested on the same dataset for various input sizes and compared initially on a low-end Laptop CPU, and then ported on two embedded platforms an Odroid device \footnote{Samsung Exynos-5422 Cortex$^{TM}$-A15 2Ghz and Cortex$^{TM}$-A7 Octa-core CPUs with Mali-T628 MP6 GPU} a Raspberry Pi3 \footnote{ Quad Core 1.2GHz Broadcom 64bit CPU} all on the same constructed aerial-view pedestrian dataset, consisting of $198$ sequential images containing $988$ pedestrians in total.

\subsection{Metrics}
The different approaches are analyzed and evaluated on the same test dataset using the following metrics:

\textbf{Sensitivity (SEN):}
This metric is defined as the proportion of true positives that are correctly identified by the detector and it is widely used as an accuracy metric, that returns the percentage of the correctly classified objects. Is calculated by taking into account the True Positives ($T^{pos}$) and False Negatives ($F^{neg}$) of the detected objects as given by (\ref{sensitivity}).

\begin{equation}
    \label{sensitivity}
    SEN = \frac{T^{pos}}{T^{pos}+F^{neg}}
\end{equation}

\textbf{Average Processing Time (APT):}
To evaluate and compare the performance for each Network, we use the average processing time metric which shows the time needed to process a single frame from a sequence of images. Specifically, this metric is the average processing time across all $N^{test\_samples}$ test images, where $t_i$ is the processing time for image $i$. 

\begin{equation}
    \label{average}
    APT = \dfrac{1}{N^{test\_samples}}\times\sum_{i=1}^{N^{test\_samples}}t_i
\end{equation}

\textbf{Average Power Consumption (APC):}
 This metric is defined as the amount of input energy (measured in watts) required for processing a single frame from a sequence of images for a particular platform. It is calculated as the summation of the power consumption at each frame devided by the total number of test images in a particular test set, where $p_i$
 is the energy consumption for image $i$.

\begin{equation}
    \label{average}
    APC = \dfrac{1}{N^{test\_samples}}\times\sum_{i=1}^{N^{test\_samples}}p_i
\end{equation}

\subsection{Evaluation of $EdgeNet$ Framework} \label{analysis}
\begin{figure}[t]
\centering
\includegraphics[width=1\linewidth]{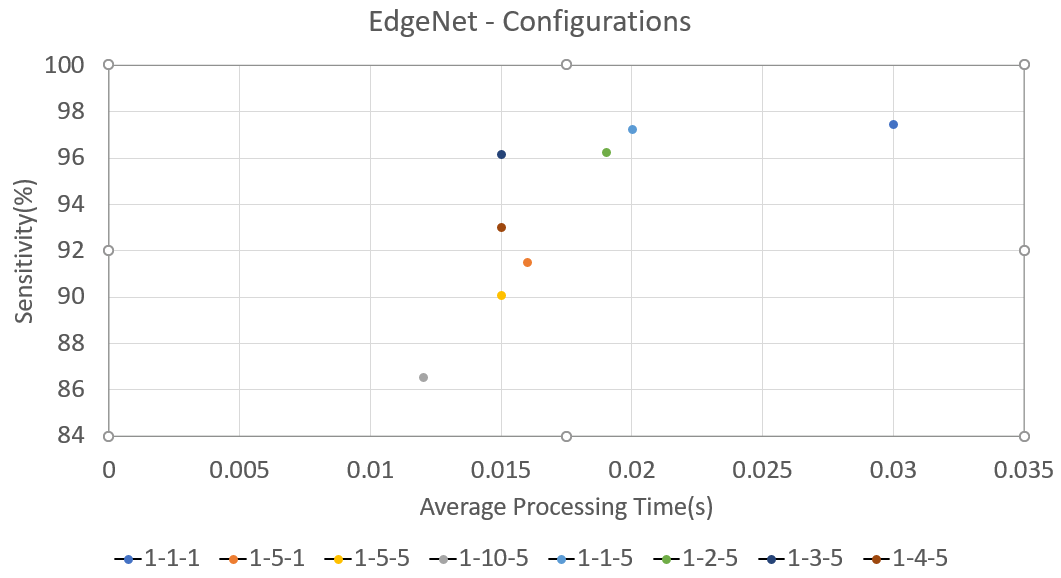}
\caption{Comparison of average processing time (CPU) and sensitivity between different $EdgeNet$ configurations for different time frames for each stage}
\label{frame}
\end{figure}

We investigate the impact of each stage for different configurations on the overall performance and accuracy. To make the framework more suitable for real-time processing at the edge and considering that stage $1$ is the most time consuming component of the framework we set its time allocation to $1$ frame. In the analysis we only vary the time allocation for stages $2$ and $3$. The average processing time and sensitivity on our constructed pedestrian test set is presented for different configurations in Fig.\ref{frame}. 

This figure shows that the time allocated at each stage has a significant impact on both performance and sensitivity of the detection framework. By increasing the time spend at stage $3$ we observe that there is a significant impact on the performance of the framework, from $0.03s$ to $0.02s$ with no impact on the sensitivity of the framework. Moreover, an increase of stage $2$ leads to a decrease for the processing time from $0.03$ to $0.015$ but at the same time there is a decrease on the sensitivity from $97\%$ to $90\%$.   Comparing the two extreme configurations $EdgeNet{{1}-{1}-{1}}$ and $EdgeNet{{1}-{10}-{5}}$ it is first observed that there is a significant variation both in terms of processing time and sensitivity. By increasing the time allocated on both stages $2$ and $3$, from $2$ to $15$ there is a significant decrease for the processing time, from $0.03$ to $0.012$ since we delay the use of the slower $DroNet\_V3$ network for a window of $15$ frames. On the other hand, since stages $2$ and $3$ operate on initial target position estimates from stage $1$ they are more susceptible to missing newly entered objects in the field-of-view. This is reflected by a decrease in sensitivity from $97\%$ to $86\%$.

\begin{figure}[t]
\centering
\includegraphics[width=1\linewidth]{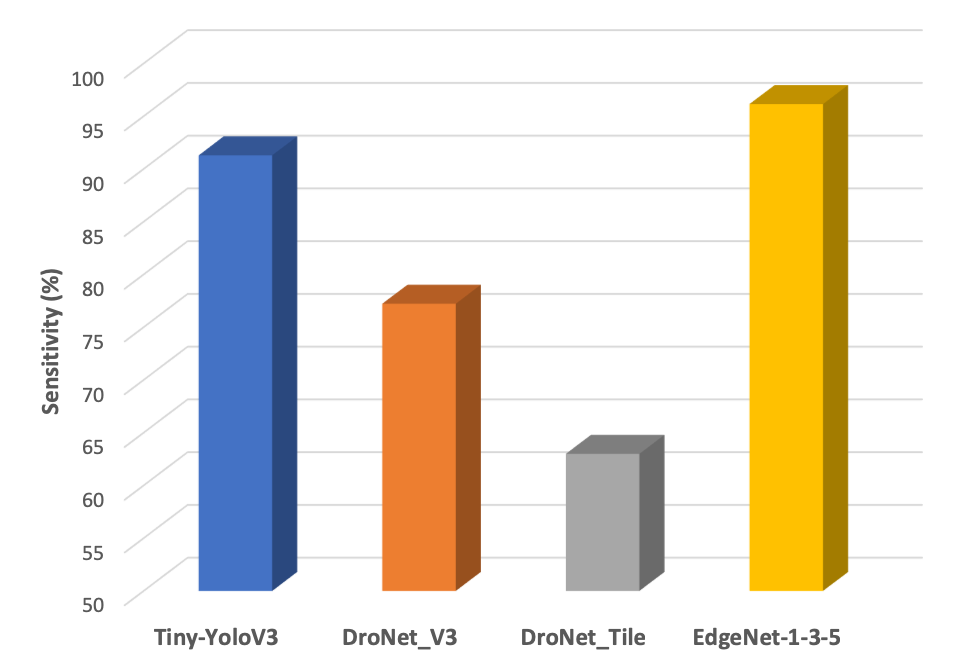}
\caption{Sensitivity of $Tiny-YoloV3$, $DroNet\_V3$, $DroNet\_Tile$ and $EdgeNet$ on different platforms}
\label{sens}
\end{figure}

This indicates that it is important to choose the appropriate values for each stage in order to avoid operating with outdated information which can lead to a reduction in accuracy. To this end it is worth exploring the design-space in between the two extremes. Our objective is to obtain the higher possible processing speed with the highest possible accuracy. As seen in Fig. \ref{frame}, the $4$ left-most points provide the best processing time. From these there is a point where the framework achieves both utilize high accuracy of $\sim97\%$ and low average processing time of $0.015$. Consequently we select $EdgeNet{{1}-{3}-{5}}$ as the best configuration in order to compare it with the different alternatives and implement it on the different edge platforms.

\subsection{Performance analysis on CPU, Odroid and Raspberry platforms}
In this section, we present an evaluation of $EdgeNet$ on different platforms, compared to the other three single-shot CNN approaches, $DroNet\_Tile$ , $DroNet\_V3$, and $Tiny-YoloV3$ with respect to sensitivity, average processing time, and energy consumption. Fig. \ref{sens} shows the sensitivity of each CNN detector on the pedestrians dataset. Sensitivity comparison, shows that $EdgeNet$ manages to keep the accuracy close to $96\%$ compared to the other CNNs, with $77\%$ for $DroNet\_V3$ and $63\%$ for $DroNet\_Tile$. This can be attributed to the fact that the single shot models resize the image prior to processing and as a result reduce the object resolution as well, leading to accuracy degradation. Even comparing with a deeper and larger CNN, $EdgeNet$ manages to outperform $Tiny-YoloV3$ by $6\%$ an indication of how well the selection process works along with the tiling approach. $EdgeNet$ spends most of its time working on image parts of the higher resolution image and as a result manages to improve accuracy by $20\%$. This shows that even though $EdgeNet$ utilizes smaller, theoretically less capable CNNs, the appropriate combination of a single deep network with the tiling for attention focusing and the tracking for fast position estimation can significantly boost accuracy.

\begin{figure}[t]
\centering
\includegraphics[width=1\linewidth]{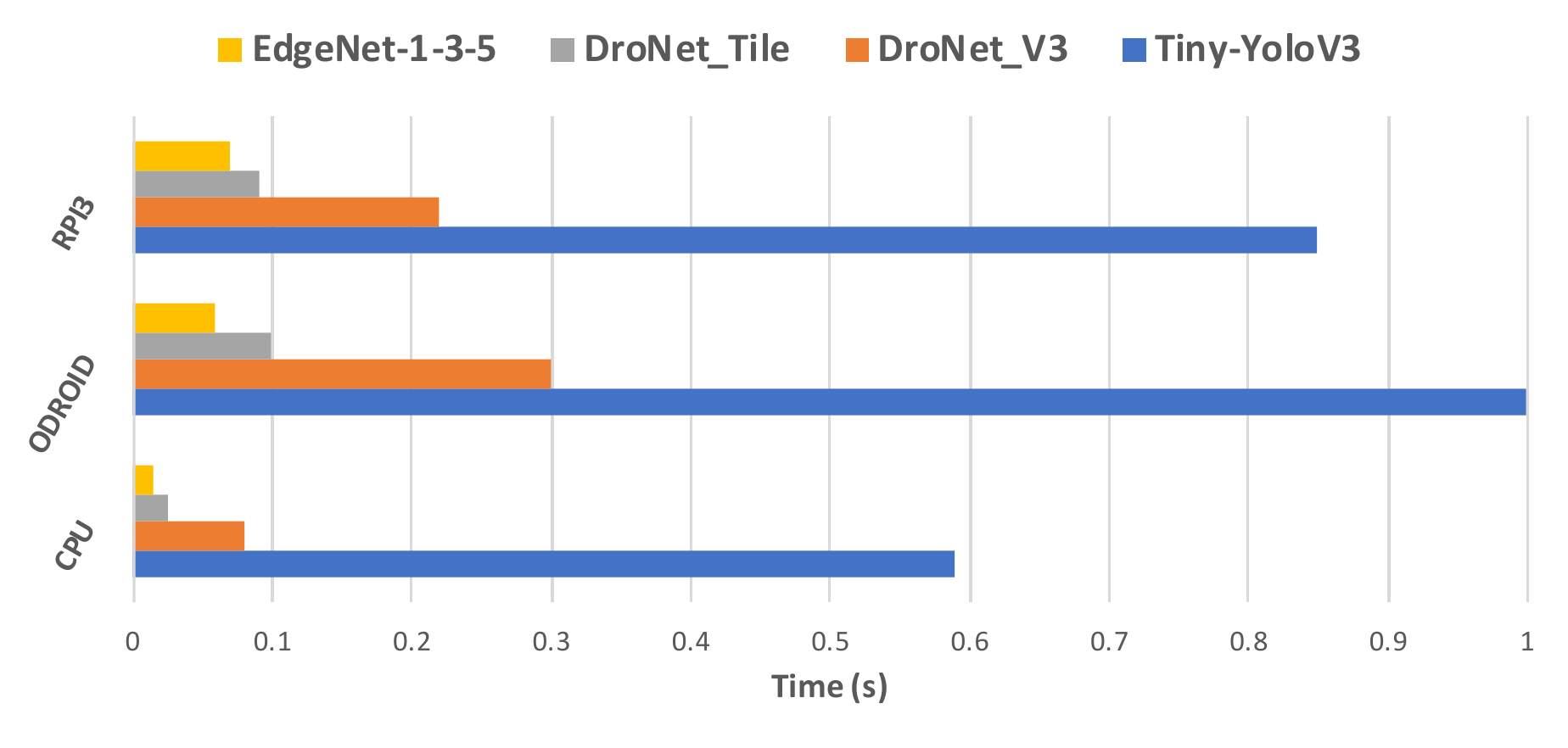}
\caption{Average Processing Time of $Tiny-YoloV3$, $DroNet\_V3$, $DroNet\_Tile$ and $EdgeNet$ on different platforms }
\label{average}
\end{figure}

\begin{figure}[h]
\centering
\includegraphics[width=1\linewidth]{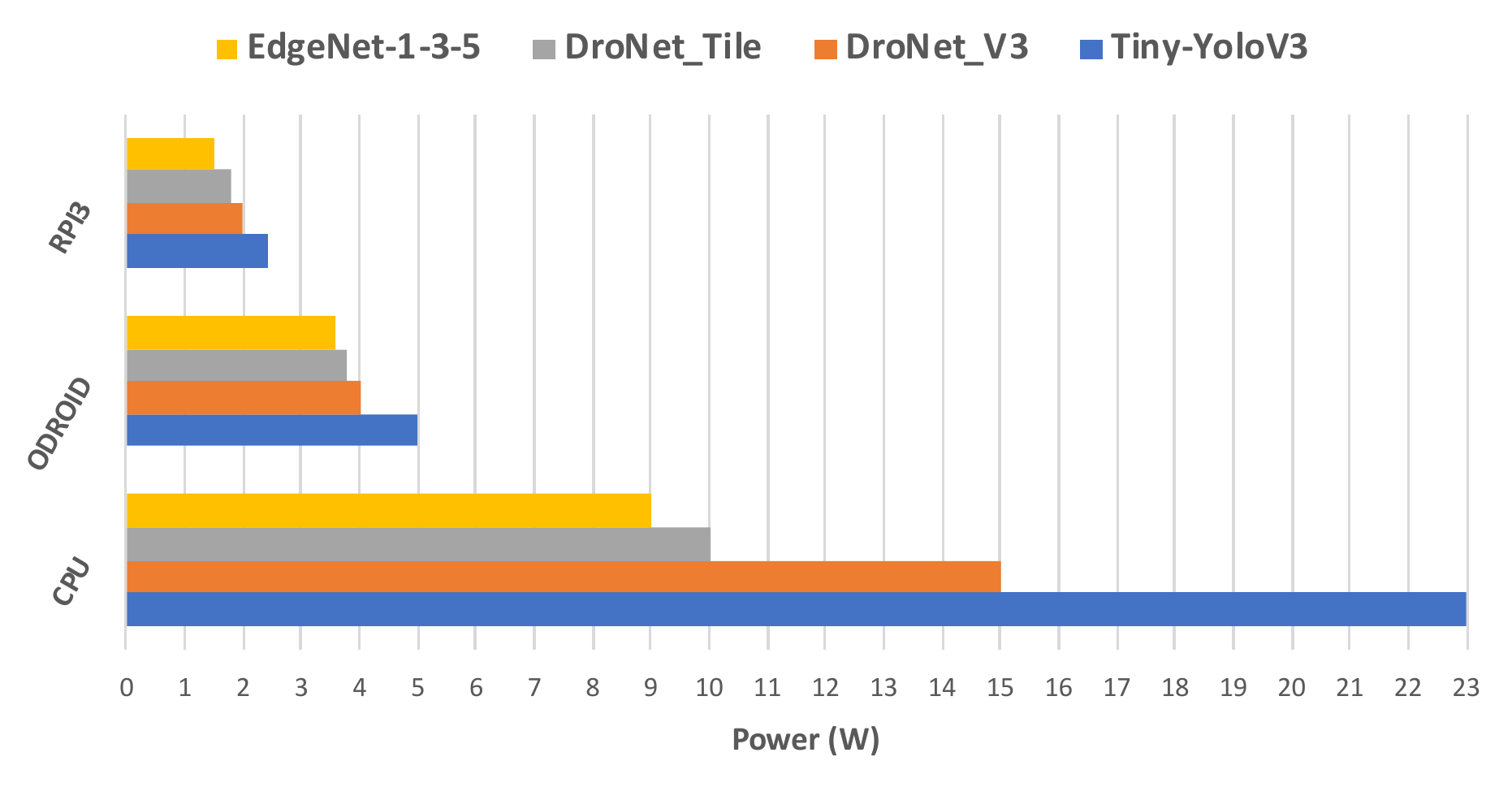}
\caption{Average Power Consumption of $Tiny-YoloV3$, $DroNet\_V3$, $DroNet\_Tile$ and $EdgeNet$ on different platforms}
\label{energy}
\end{figure}

\begin{figure*}
\centering
\includegraphics[width=\textwidth]{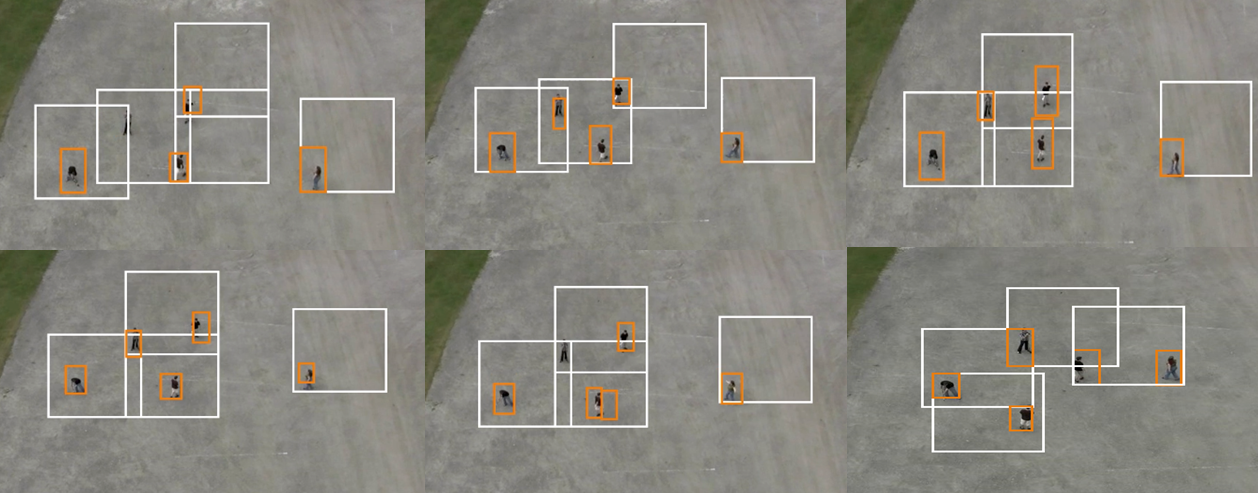}
\caption{$EdgeNet$ selection process on different frames of the test set. White boxes indicates the proposed tiles for processing and orange is the actual detection of the objects}
\label{edgenet}
\end{figure*}

Fig. \ref{average} shows the average processing time on different evaluation platforms. First, it is noticeable that the selected configuration of $EdgeNet$ is faster on all platforms than the other approaches. In all devices there is a reduction of the APT from $~60-100 \times$, which shows that with no impact, and in some cases an increase of sensitivity, $EdgeNet$ manages to boost the inference time of the detector. It is also worth noting that the performance of $EdgeNet$ adapts to the activity (number and location of pedestrians) in the scene due to its dynamic nature, whereas the processing time of the other approaches is constant regardless of the frame content. Fig. \ref{edgenet} shows different time instances of the selection of tiles and the detections on the constructed dataset, another example on the way $EdgeNet$ can select different tiles for processing and at the same time cover all the objects efficiently. Overall, $EdgeNet$ is able to run in real-time on all platforms with an average processing time between $0.02 - 0.06$ on all devices.  As such, it verifies our claim that an intelligent processing pipeline can be more efficient than a single CNN, for use in mobile/edge devices.

Moreover, as shown in Fig. \ref{energy}, $EdgeNet$ leads to a reduction of the average power consumption which makes it the most power efficient detector compared to the other CNNs. The reduction of processed data along with the use of CNNs with small input size and the tracker, has a direct impact on the average power consumption on all platforms due to the reduction of computation. Comparing to $Tiny-YoloV3$, which consumes the most power compared to all the other networks, there is a $14W$ decrease of the power consumption on the CPU platform and a decrease of $1-1.5W$ on the other two platforms.

\section{Conclusion \& Future Work}\label{conc}
This paper proposed a three-stage framework for a more efficient object detection on higher resolution images for edge/mobile devices. We have demonstrated that an intelligent data reduction mechanism can go a long way towards improving the overall accuracy and focus the computation on the important image regions. Furthermore we have shown that by selectively choosing the best CNNs to use based on the position and proximity of targets between them there are significant benefits in terms of performance and accuracy. Overall, $EdgeNet$ manages to provide promising performance between $15 - 50$ frames-per-second, with $96\%$ accuracy and a power consumption between $1.5 - 9W$, depending on the inference device. Future research plans include the optimization, using binarization and pruning techniques, of each individual CNN to further improve the speed. Moreover, we plan to test EdgeNet on various scenes and different conditions and incorporate the movement of the objects in the selection process in order further increase the detection accuracy in case of high movement of both the objects and the camera.






\section*{Acknowledgement}
Christos Kyrkou gratefully acknowledges the support of NVIDIA Corporation with the donation of the Titan Xp GPU used for this research.

\bibliographystyle{ACM-Reference-Format}
\balance
\bibliography{references.bib}

\end{document}